\pdfoutput=1
\documentclass[conference]{IEEEtran}
\IEEEoverridecommandlockouts
\usepackage{cite}
\usepackage{amsmath,amssymb,amsfonts}
\usepackage{algorithmic}
\usepackage{graphicx}
\usepackage{textcomp}
\usepackage{xcolor}
\usepackage{amsmath,amsfonts}
\usepackage{algorithmic}
\usepackage{algorithm}
\usepackage{array}
\usepackage{textcomp}
\usepackage{stfloats}
\usepackage{url}
\usepackage{verbatim}
\usepackage{graphicx}
\usepackage{caption}
\usepackage{subcaption}

\usepackage{flushend}

\usepackage{amsmath}
\usepackage{multirow}

\usepackage{enumitem}
\usepackage{cite}
\usepackage{pdfpages}
\def\BibTeX{{\rm B\kern-.05em{\sc i\kern-.025em b}\kern-.08em
    T\kern-.1667em\lower.7ex\hbox{E}\kern-.125emX}}
\begin{document}

\title{Simpler is better: Multilevel Abstraction with Graph Convolutional Recurrent Neural Network Cells for Traffic Prediction
}

\author{Naghmeh Shafiee Roudbari, Zachary Patterson, Ursula Eicker, Charalambos Poullis\\
    \textit{Gina Cody School of Engineering and Computer Science} \\
    \textit{Concordia University}\\
    \textit{Montreal, QC, Canada}
    }


\maketitle

\begin{abstract}
In recent years, graph neural networks (GNNs) combined with variants of recurrent neural networks (RNNs) have reached state-of-the-art performance in spatiotemporal forecasting tasks. This is particularly the case for traffic forecasting, where GNN models use the graph structure of road networks to account for spatial correlation between links and nodes. Recent solutions are either based on complex graph operations or avoiding predefined graphs. This paper proposes a new sequence-to-sequence architecture to extract the spatiotemporal correlation at multiple levels of abstraction using GNN-RNN cells with sparse architecture to decrease training time compared to more complex designs. Encoding the same input sequence through multiple encoders, with an incremental increase in encoder layers, enables the network to learn general and detailed information through multilevel abstraction. We further present a new benchmark dataset of street-level segment traffic data from Montreal, Canada. Unlike highways, urban road segments are cyclic and characterized by complicated spatial dependencies. Experimental results on the METR-LA benchmark highway and our MSLTD street-level segment datasets demonstrate that our model improves performance by more than 7\% for one-hour prediction compared to the baseline methods while reducing computing resource requirements by more than half compared to other competing methods. 
\end{abstract}


\section{Introduction}
Spatiotemporal forecasting is an essential tool for understanding variations in space-time data and ultimately helps inform resource allocation, risk management, and policy-making decisions. It has long been a popular field of research in machine learning. Even though many influential machine learning (ML) approaches have been proposed in this field, there is still a considerable gap between the state-of-the-art and accurate predictions, particularly when it comes to traffic forecasting. 


Traffic behaviour on more granular urban street networks is fundamentally different from highways due to differences in spatial complexity and temporal variability:

\paragraph{Spatial complexity differences} 
Connectivity is the density of connections in a road network \cite{TDMEncyclopedia}. Urban streets include numerous short links and intersections, so the density of connections in the urban road network is way higher than those in the highway network and they are characterized by more complex spatial dependencies than highways.

\begin{figure*}[!t]
    \centering
    \begin{subfigure}{0.32\textwidth}
        \includegraphics[width=\textwidth]{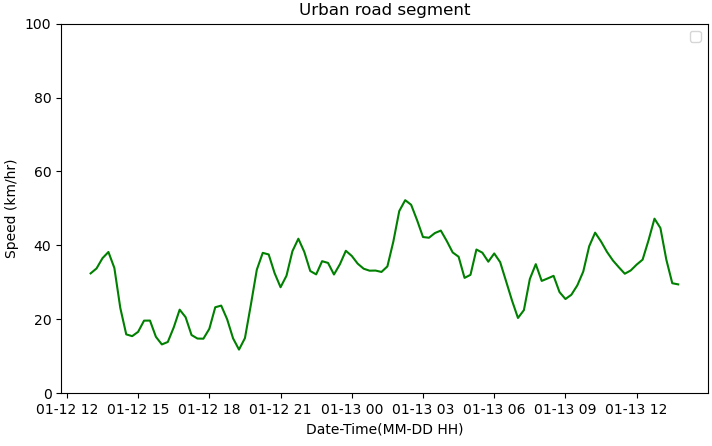}
        \caption{}
        \label{subfig:Urban_green}
    \end{subfigure}
    \hfill
    \begin{subfigure}{0.32\textwidth}
        \includegraphics[width=\textwidth]{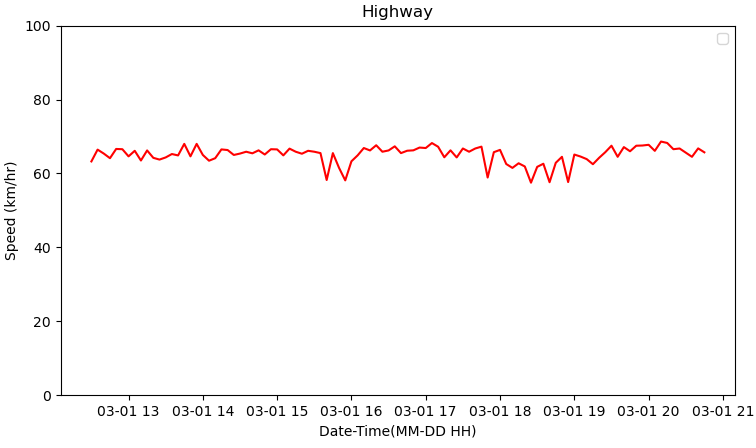}
        \caption{}
        \label{subfig:highway_red}
    \end{subfigure}
    \hfill
    \begin{subfigure}{0.32\textwidth}
        \includegraphics[width=\textwidth]{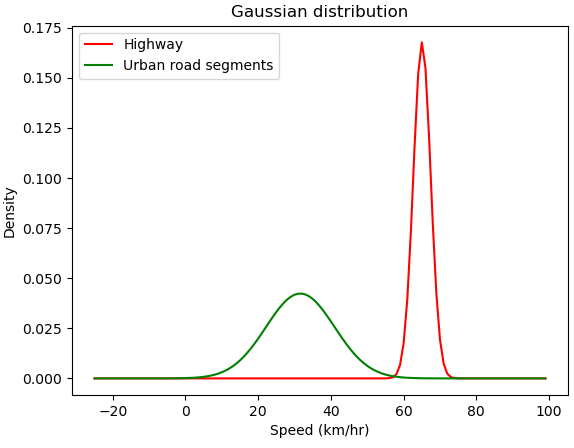}
        \caption{}
        \label{subfig:gaussian}
    \end{subfigure}
    \caption{Differences in data variability in two different context of traffic. Traffic speed changes for 24 hours (a) An urban road segment from our new MSLTD street-level dataset, and (b) a Highway sensor from the METR-LA dataset (c) Gaussian distribution of the speed in (a) (green) and (b) (red). }
    \label{fig:temporal_plot}
\end{figure*}

\paragraph {Temporal variability differences} Because of the impact of traffic signals and traffic volume, the speed variability on an urban road segment is significantly higher compared to a highway \cite{Karr2002d}. Fig. \ref{fig:temporal_plot} explores the difference in speed variability between the two different contexts of traffic. Fig. \ref{subfig:gaussian} shows the Gaussian distributions for speed in an urban road (\ref{subfig:Urban_green}) and highway (\ref{subfig:highway_red}). In urban road segments, traffic speed is lower than  highway speed. Besides, for urban road segment data, the Gaussian graph is short and wide, proving the standard deviation is significant compared to highway speed data with a narrow and tall Gaussian distribution.


\IEEEpubidadjcol


In this paper, we present experiments on urban streets and highway benchmark datasets that demonstrate the robustness of our proposed model against substantial temporal variability and the ability to learn more complex spatial correlations than baseline methods. 

Previous research has studied different approaches such as statistical models, machine learning techniques, and deep learning methods to explore the spatial and temporal features in traffic data. RNN family \cite{hochreiter1997long,chung2014empirical} combined with GNNs \cite{wu2020comprehensive} have shown great potential to learn complexities in traffic data effectively. Even though the GNN-RNN based methods introduced so far address both the spatial and temporal dependencies, these methods either do not fully exploit the spatial information or have an over-complex structure, particularly in graph convolution operation, that affects the training time. 
This paper addresses the problem of cyclic graph networks with complex spatiotemporal inter-dependencies. The main contributions of this paper include:

\begin{enumerate}
    \item Multilevel encoder architecture: We formulate the problem as a sequence-to-sequence modelling task and introduce a novel multilevel encoder architecture that abstracts and combines dependencies at multiple complexity levels. Uniquely, the proposed technique handles sparse and dense road networks and enhances the accuracy of the predictions.
    
    \item Sparse architecture: Furthermore, when compared with state-of-the-art, our method employs a sparse architecture that improves time efficiency by reducing the computational complexity of the training. 

    \item We present our experiments on the application of traffic forecasting and report the results on benchmark datasets of highway-level data METR-LA and the newly released street-level MSLTD dataset presented in this paper. Detailed descriptions of the datasets are provided in Section \ref{dataset}.
\end{enumerate}

The rest of the paper is organized as follows. Section \ref{sec:literature} reviews the current state of the literature in traffic flow forecasting. Section \ref{sec:model} formally introduces the model, and Section \ref{sec:methodology} addresses the traffic prediction problem using the proposed model. Finally, Section \ref{sec:experiments} shows numerical experiments on real-world datasets and compares the proposed model's results with other state-of-the-art models reported in the literature.

\section{Related Work}
\label{sec:literature}
A wide range of data-driven methods regarding traffic prediction has been proposed in the past. Traditional approaches for predicting traffic state include Kalman-filter based methods \cite{guo2014adaptive,xie2007short}, Vector Auto Regressive models \cite{chandra2009predictions}, Autoregressive Integrated Moving Average (ARIMA) \cite{van1996combining} and variants such as seasonal ARIMA \cite{kumar2015short} and Space-Time ARIMA \cite{ding2011forecasting}. These statistical methods assume that traffic data has a fixed variability pattern. In reality, however, traffic data does not obey stationary assumptions, so traditional approaches mostly fail to achieve a good performance. To overcome this issue, machine learning methods with promising performance compared to linear methods became popular in time-series forecasting. Since then, many machine learning models, including K-nearest neighbour approaches \cite{cai2016spatiotemporal}, Support Vector Machine models \cite{yao2006research,cong2016traffic}, and Bayesian Network approaches \cite{sun2006bayesian} have been applied to traffic forecasting problem. 

Before the emergence of deep learning, various shallow Neural network (NN) models were proposed to address traffic prediction, such as Multi-Layer Perceptron networks (MLP) \cite{innamaa2000short} to predict traffic parameters such as speed and flow and study the effect of MLP parameters on prediction. Similarly, radial basis function (RBF) neural networks have been used for freeway traffic flow prediction with fuzzy c-means clustering to find the center position of the hidden layer \cite{xiao2004study}. Although shallow NN models improved traffic prediction compared to traditional methods, they are still limited in their ability to capture spatial and temporal dependencies in complex traffic data. This inspired researchers to apply deep learning methods. In \cite{lv2014traffic}, the authors used Stacked Auto Encoders (SAE) with a standard logistic regression model to train the network in a supervised manner for traffic flow prediction.  \cite{yang2016optimized} proposed another class of autoencoders named stacked autoencoder Levenberg-Marquardt model to improve forecasting accuracy by learning features through multiple layers in a ``greedy'' way. \cite{huang2014deep} proposed a deep architecture comprising two main modules, a deep belief network (DBN) to learn the features automatically and a multitask regression layer. 

A Recurrent Neural Network (RNN) is a type of neural network containing at least one loop within the structure of network connections, enabling RNNs to have internal memory to remember a summarization of previous states and understand sequential information. This feature makes RNN models an excellent building block for a time-series forecasting network. LSTM and GRU are gated RNNs designed to overcome the vanishing gradient problems. Gated RNNs can also successfully learn temporal correlation in traffic data by capturing long-term information. Many RNN-based models such as bidirectional LSTM \cite{cui2018deep}, Mixture Deep LSTM \cite{yu2017deep}, shared hidden LSTM \cite{song2016deeptransport} models and GRU models \cite{agarap2018neural} have been applied to traffic forecasting. \cite{tian2015predicting} proposed an LSTM RNN model and demonstrated how it outperforms previously proposed models, including SVM, single layer feed-forward neural network (FFNN), and stacked autoencoder (SAE). \cite{liu2017short} uses an LSTM model for travel time prediction. \cite{ji2017forecast} has applied GRU models to predict bus trip demand and shown how GRU and NN models outperform ARIMA models for different prediction horizons. Although RNN-based methods consider the dynamic variations of traffic conditions, the future traffic state is affected by spatial dependencies besides the temporal changes. Since these models fail to explore spatial complexities, they cannot generate accurate predictions.

Recent studies \cite{wu2016short}, \cite{ma2017learning} have attempted to model spatial dependencies using the Convolutional Neural Networks (CNN) to address this issue. CNN models are a class of neural networks used primarily for extracting features from images \cite{oquab2014learning}. To formulate the traffic prediction problem using CNNs, \cite{ma2017learning} constructed a time-space matrix, where each element in the matrix contains information about the time interval and the street section associated with that element; \cite{yu2017spatiotemporal} also proposed CNN models to investigate the spatial correlation in traffic data; \cite{lv2018lc} proposed a combination of RNN and CNN models to take advantage of both by learning time-series dependencies and capturing the features related to the road network; and \cite{guo2019deep} introduced 3D convolution to capture spatial correlations and temporal correlations from both long-term patterns and local patterns. However, CNN models can only extract spatial dependencies from a rigid grid structure representing the information, and abstract features extracted by CNN represent the relationship in Euclidean space. 

There are numerous applications where Euclidean space cannot perfectly describe spatial dependence such as chemical synthesis, social network analysis, 3D vision and transportation networks. A graph is a powerful data structure that can express complex relationships in unstructured data \cite{wu2020comprehensive}. Graph Neural Networks (GNNs) are deep learning-based methods that have been widely applied to graph domains in recent years. Neural networks based on the graph theory \cite{butler2006spectral}, \cite{bruna2013spectral} process the graph-structured data. In order to capture spatial dependence in a graph context, different types of Graph Convolutional Network (GCN) models have been introduced. GCNs are categorized into two main domains, spectral and spatial. The spectral graph convolution is based on graph Laplacian to localize a graph signal on the spectral domain, and the spatial graph convolution is the aggregation of information within a node's neighbourhood \cite{zhang2019graph}. Most recent studies \cite{li2017diffusion}, \cite{cui2019traffic}, \cite{bai2020adaptive} on traffic forecasting focus on using a hybrid architecture of GCN to extract the spatial dependencies and RNN to explore the temporal changes. \cite{li2017diffusion} maps traffic state to a diffusion process on a directed graph and uses diffusion convolution integrated with gated recurrent units to capture both spatial and temporal dependencies. A sequence-to-sequence model is used in this study as a learning structure, in which they employ a scheduled sampling approach for training. \cite{zhao2019t} proposes a temporal graph convolution for traffic prediction and uses GRU to understand temporal dynamics. \cite{cui2019traffic} presents a traffic graph convolution operator and a convolutional LSTM to predict traffic speed. \cite{wu2020connecting} avoid using a pre-defined graph in their proposed model, which includes a graph learning layer, a graph convolution component, and a temporal convolution component. \cite{bai2020adaptive} also argues against pre-defined graphs since domain knowledge is necessary to generate the graph, as such information expressed by the graph might not be directly related to the prediction task, and may even introduce bias.

Sequence-to-sequence architecture is an end-to-end approach to processing sequence data. It includes a wide range of applications like speech modelling \cite{sutskever2014sequence}, natural language processing \cite{ma2016end}, and recently time-series forecasting to enhance capturing long-term dependencies and enabling multistep prediction. \cite{li2017diffusion} used sequence-to-sequence modelling for traffic speed prediction by applying an encoder-decoder architecture, where the encoder and decoder blocks consisted of their proposed diffusion convolution recurrent neural network. \cite{zheng2020gman} proposed a sequence-to-sequence architecture consisting of attention blocks to predict traffic. \cite{wang2020long} proposed an encoder-decoder design with LSTM blocks in addition to a control layer on top of it to cumulatively learn the data. 

The literature indicates the strong potential of RNN models to learn the temporal information in traffic data. GCN has also achieved satisfactory results in unlocking the spatial relation in the traffic network. Although these approaches have improved accuracy considerably compared to previous methods, they either employ a complex network design that increases the number of parameters and training time or cannot achieve comparable accuracy to other deep learning tasks. This paper proposes a novel multilevel sequence-to-sequence architecture to extract discriminative features effectively. The proposed model is based on a sparse architecture graph convolutional GRU enabling the network to reduce computational costs.

\section{Proposed Model}
\label{sec:model}
\subsection{Graph Notation}
A graph is a data structure with a solid mathematical basis used to represent complex interactions between a set of objects. A graph denoted by $\mathcal{G(V, E)}$ includes a set of $n \in \mathbb{Z}$ nodes represented as $\mathcal{V}= \{v_1, v_2, \dots, v_n \}$ and a set of edges $\mathcal{E}=\{e_{i,j} | i,j \in \{1,2,..,n\}\}$, where each edge $e_{ij}$ indicates an interaction between nodes $i$ and $j$. An adjacency matrix $\textbf{A}_{n\times n}$ is one way to represent a graph. $\textbf{A}$ is a square matrix, where a non-zero element at $\textbf{A}_{i,j}$ indicates the existence of an edge connecting nodes $i$ and $j$. Every node of a graph can have a set of features/attributes associated with it. To this end, a matrix $\textbf{X} \in \mathbb{R}^{n\times d}$ denotes an attribute representation of graph G, where each row of the matrix is a vector that corresponds to a specific node of the graph representing its $d$ features \cite{hamilton2020graph}. 

In a spatiotemporal graph, the objects that are the resource of temporal changes are considered nodes. The feature matrix denoted as $\textbf{X}^t \in \mathbb{R}^{n\times d}$ dynamically changes over time $t$ \cite{wu2020comprehensive}. To represent a spatiotemporal forecasting problem as a graph, one must first translate the temporal data sequence to a graph signal. Each node $v_i$ is assigned a feature vector $X_i^t \in \mathbb{R}^{1\times d}$ where $d$ is the length of the historical time-series of node $i$ at time $t$ needed to predict the future sequence.

\subsection{Problem statement}
Spatiotemporal forecasting aims to predict the future sequence of a parameter given its historical time-series. More formally, this is equivalent to finding a function $f: X^{t}, A \rightarrow Y^{t+h}$ which given an adjacency matrix $\textbf{A}$ representing the graph structure and a sequence of observed data $X^{t}$ representing $d$ previous historical states of $n$ nodes,

\begin{figure*}[t]
    \centering
    \includegraphics[width=\textwidth]{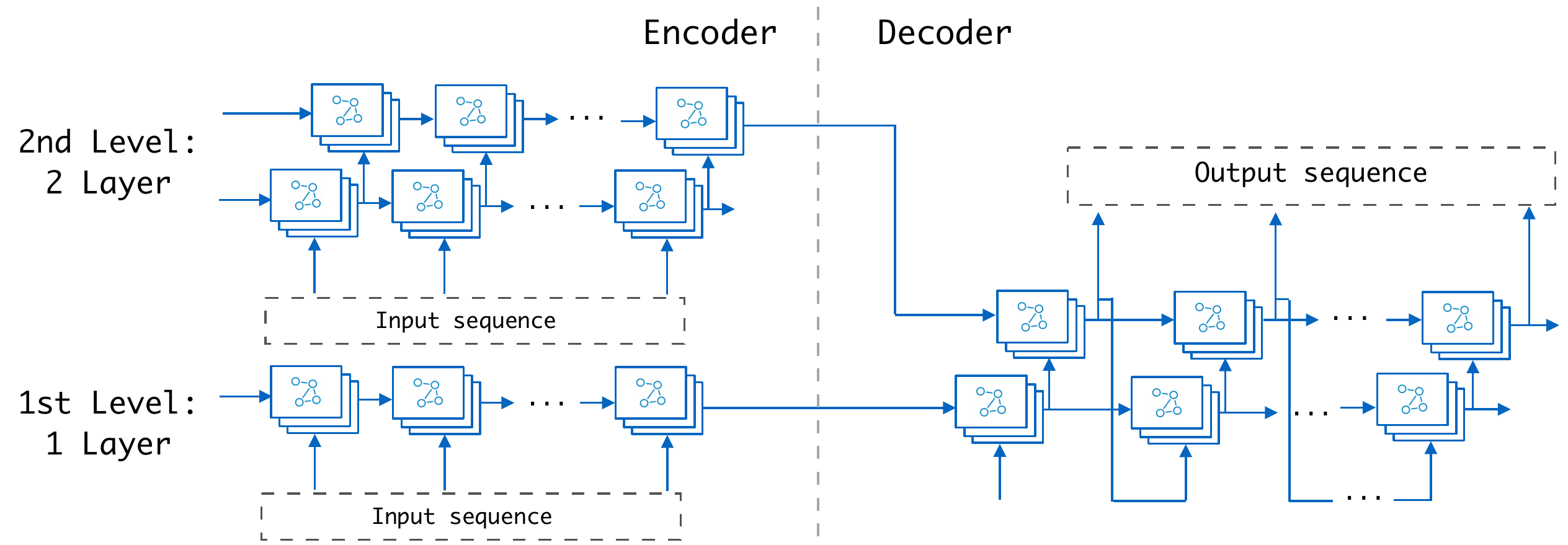}
    \caption{Multilevel Sequence-to-Sequence (MLS2S) Architecture for a 2-level encoder. The output from the last layer of each encoder is used to initialize the decoder. The number of encoder levels equals number of decoder layers.}
\label{fig:architecture}
\end{figure*}

\begin{center}
\vspace{.3cm}
$X^t =
\begin{bmatrix}
x_0^{t-d+1} & \dots & x_0^{t-1} & x_0^{t} \\
x_1^{t-d+1} & \dots & x_1^{t-1} & x_1^{t} \\
\dots & \dots & \dots & \dots \\
x_{n-1}^{t-d+1} & \dots & x_{n-1}^{t-1} & x_{n-1}^{t} \\
\end{bmatrix}_{n*d}$
\end{center}
\vspace{.3cm}
it gives a sequence of values $Y$ at a future time $t+h$ of up to $h$ steps. Thus, $Y^{t+h}$ is given by, 
\vspace{.3cm}
\begin{center}
$Y^{t+h} =
\begin{bmatrix}
x_0^{t+1} & x_0^{t+2}& \dots & x_0^{t+h} \\
x_1^{t+1} & x_1^{t+2}& \dots & x_1^{t+h} \\
\dots & \dots & \dots & \dots \\
x_{n-1}^{t+1} & x_{n-1}^{t+2}& \dots & x_{n-1}^{t+h} \\
\end{bmatrix}_{n*h}$
\end{center}
\vspace{.3cm}

In this context, the graph nodes $\mathcal{V}$ represent a set of $n$ objects generating time-series data, and the existence of an edge $e_{ij}$ denotes a dependency between objects $i$ and $j$. Since the rows of the matrices above correspond to the graph's nodes, the prediction is network-wide.

\subsection{Cell Architecture}
\label{cell}
Cells are the basic building blocks of the prediction network. A cell consists of two main components. First, the spatial module captures the spatial features given the road network graph. The second component combines the spatial module operation with RNNs to learn temporal dependencies. As explained in the following sections, our work focuses on finding a practical sequence-to-sequence model solution, which at the same time maintains a simple cell design. As a result, the computational complexity for training decreases by more than a factor of 2 compared to state-of-the-art employing more complex architectures. 

\paragraph{Spatial Learning Module}
Graph Convolutional Network (GCN) models aggregate node features and graph structure information to capture spatial correlation in data. For the spatial module, we leverage the spectral graph convolution operation based on normalized Laplacian by utilizing the simplified operation using Chebyshev polynomial approximation \cite{kipf2016semi}:
\begin{equation}
\label{eq:graph_conv}
{H}^{(k)}= \sigma (\hat{D}^{-1/2}\hat{A}\hat{D}^{-1/2}H^{(k-1)}W^{(k)})+ b^k
\end{equation}
\begin{center}
$\hat{A} = A + I$,
$\hat{D} = D + I$
\end{center}
where $A$ is the adjacency matrix of graph $\mathcal{G}$, $D$ is the degree matrix, a diagonal matrix $n*n$ where $D_{ii}$ represents the number of edges connected to node $i$.  $D_{ii}$  element value can be acquired by summing up the row or column elements associated with node $i$ in an undirected graph's adjacency matrix. $W^k$ and $b^k$ are the trainable weight and bias matrices for layer $k$, and $\sigma$ is the activation function. $H^{k-1}$ and $H^k$ are the input signal and output of GCN operation on layer $k$, respectively.

\paragraph{Temporal Learning Module}
RNNs have achieved promising performance for natural language processing and sequence modelling applications. RNN gated models like Gated Recurrent Units (GRUs) can account for long-range dependencies based on their built-in architecture. We use GRU layers in our temporal module design to extract the temporal dynamics.

We integrate the spatial and temporal learning modules by replacing all the fully-connected layers of the GRU with the GCN layer introduced in the spatial module leading to, 
\begin{equation}
\label{eq:rt}
r^{t}= \sigma (\mathcal{G}_{conv}(A,[X^t,h^{t-1}])+ b_r
\end{equation}
\begin{equation}
\label{eq:ut}
u^{t}= \sigma (\mathcal{G}_{conv}(A,[X^t,h^{t-1}])+ b_u
\end{equation}
\begin{equation}
\label{eq:ct}
c^{t}= tanh (\mathcal{G}_{conv}(A,[X^t,r^t *h^{t-1}])+ b_c
\end{equation}
\begin{equation}
\label{eq:ht}
h^{t}= (u^t * h^{t-1})(1.0-u^t)*c
\end{equation}
where $\mathcal{G}_{conv}$ is the graph convolution operation from Equation \ref{eq:graph_conv}, $r^t$ and $u^t$ are the reset gate and the update gate, $c^t$ is the cell state, $X^t$ is the input signal, $\sigma(.)$ and $tanh(.)$ are activation functions, and $h$ is the hidden state. We aim at an efficient solution with reduced training time; hence we avoid adding complexity to the cell design by keeping the architecture simple. 

\subsection{Multilevel Sequence-to-Sequence Architecture (MLS2S)}
Generally, Encoder-Decoder architectures include a pair of networks trained in tandem on target sequences given their input sequences. The encoder transforms the input series with variable size to a hidden state with a fixed size, and the decoder converts back the hidden state representation to a variable size prediction series \cite{cho2014learning}. It is not necessary to keep the input and output sequences fixed.

In this paper, we propose a novel architecture that modifies the original encoder-decoder structure and uses multilevel encoders. The multilevel encoder integrates simple abstraction from the first level with the more detailed abstraction of higher levels and constructs a stack of hidden states with expressive information to initialize the decoder layers. The number of encoder levels matches that of the decoder layers since the final output state of every level of encoder is used as the input of the decoder layer at the same level. An example two-level MLS2S is shown in Fig. \ref{fig:architecture}. The input sequence is a fixed-length time-series of historical data that indicates a sequence of inputs. The same input sequence feeds into the encoders at all levels. 

The decoder uses its previous hidden state as input and the encoder output as its initial state to generate the prediction. For each encoder level, only the final layer hidden state is used as the initial state of the corresponding decoder layer.

\paragraph{Multilevel Encoder} The encoders are the cell blocks introduced in Section \ref{cell}. Given each element of the input sequence, the encoder layer $l$ updates its state as follows:

\begin{equation}
\label{eq:encoder1}
h_l^t = f_l(h_{l-1}^t, h_l^{t-1},A)
\end{equation}
where $h_0^t$ equals the input sequence $X^T$. After reaching the end of the sequence, the final hidden state of the encoder is acquired. The Multilevel Encoder is a network of multiple encoders reading the same input sequence. The first level is a one-layer encoder, and the subsequent number of layers increases incrementally for higher levels. Ultimately, the final hidden state of all levels are concatenated and become the initial hidden state of the decoder. For an $L$-level encoder, the final stacked hidden state is given by,
\begin{equation}
\label{eq:encoder2}
h = Concat[h_1^t, h_2^t, ... , h_L^t]
\end{equation}

\paragraph{Decoder}: The decoders also follow the cell block architecture; they are trained to produce the predicted series given the hidden state and the previous output. For multilayer decoder, the hidden state of the decoder on layer $l$ at time $t$ is computed by:
\begin{equation}
\label{eq:encoder3}
Y_l^t = f_l(y_{l-1}^{t-1},h_l^{t-1},A)
\end{equation}
where $h_l^0$ equals the corresponding element of the final hidden state from encoder, $h[l]$. The final predicted sequence, for an $L$-layer decoder equals $Y_L$. 


\section{Methodology}
\label{sec:methodology}
We investigate the use of the MLS2S model on traffic speed forecasting. We predict traffic speed parameters given past traffic speed observations and the road network. We represent the road network's graph $\mathcal{G}$ as an unweighted adjacency matrix $A$. Graph nodes $\mathcal{V}$ represent a set of $n$ road segments or highway sensors, and a non-zero value for $e_{ij} $ denotes that the two road segments $i$ and $j$ are connected. Every element of the historical and future sequence provides traffic speed at the specific time range $t$ for all the nodes,  $\{v_1, v_2, \dots, v_n \}$ and the final prediction covers the entire road network.

At timestamp $t$, we simultaneously look ahead at the next $\eta$ steps $\{t+1, t+2, \dots, t+\eta\}$. We train the network using a Mean Absolute Error (MAE) loss between the predicted sequence and the ground truth given by,
\begin{equation}
	\label{eq:mae}
	MAE = \sum_{i=1}^{h}|Y^{t+i} - X^{t+i}|
\end{equation}
where $X^t$ is a vector providing ground truth speed values for the entire network at time $t$ and $Y^t$ is a vector providing the network-wide predicted speed values at time $t$.

We use backpropagation with Adam optimizer for the training with a learning rate decay ratio to decrease the learning rate by a constant factor as a function of time. The general intuition behind this strategy is to avoid being stuck in local minima, speed up the training process during the initial steps, and reduce the learning rate to prevent oscillation in the final epochs. We follow this approach because starting with a larger learning rate prevents the network from learning noisy data while decaying the learning rate over time enables the network to learn complex patterns \cite{you2019does}.

\begin{figure}[!t]
    \centering
    \begin{subfigure}{0.5\textwidth}
    \includegraphics[width=\textwidth]{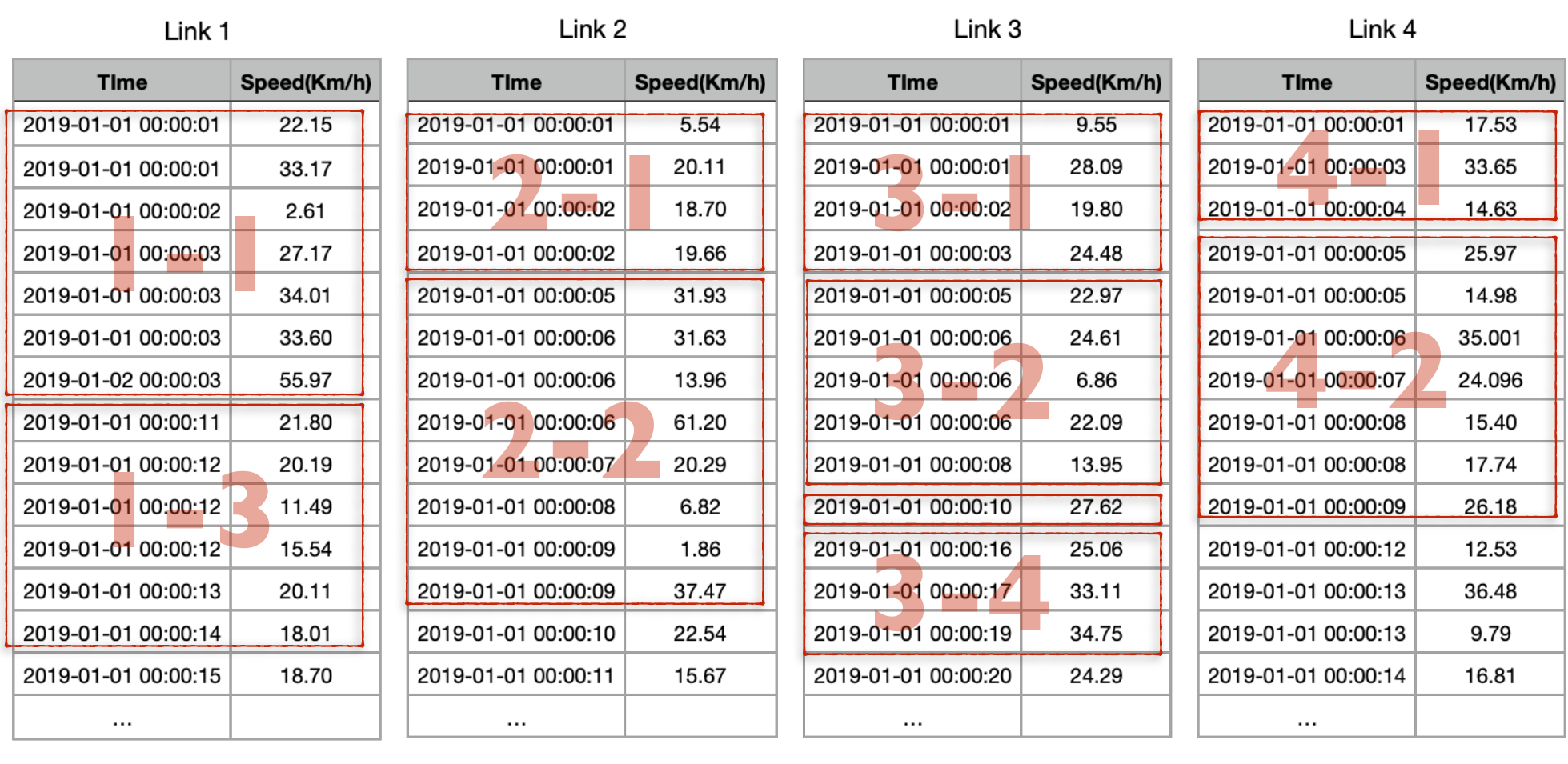}
        \caption{}
        \label{subfig:links_table}
    \end{subfigure}
    \begin{subfigure}{0.5\textwidth}
    \includegraphics[width=\textwidth]{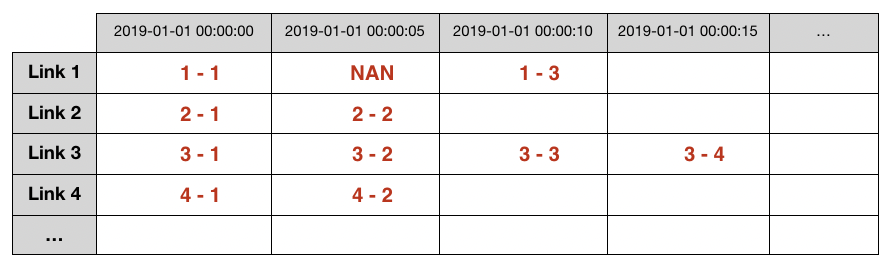}
        \caption{}
        \label{subfig:speed_matrix}
        
    \end{subfigure}
    \caption{MSLTD dataset preparation process (a) Travel time records provided for links oredered by time(b) Speed matrix table covering the whole time range in 5 minute intervals for each link (the elements labeled in red, point to the group of travel time records in corresponding link table. The blank elements are to be filled based on the rest of information from links table)}
    \label{fig:maps1}
\end{figure}

\section{Experiments}
\label{sec:experiments}
\subsection{Dataset Description}
\label{dataset}
We present experiments on a highway-scale dataset and our newly introduced street-level dataset to demonstrate the efficacy of the proposed approach. We quantitatively evaluate the MLS2S model on two real-world traffic datasets: METR-LA containing the highway data of Los Angeles, USA, and the proposed MSLTD street-level segment traffic data from Montreal, QC, Canada. 

\subsubsection{Highway benchmark dataset METR-LA}
This benchmark dataset contains the average traffic speed at 5-minute intervals for 207 locations. The data is collected from sensors located on highways in Los Angeles, USA, over four months from March 2012 to June 2012 \cite{jagadish2014big}. The coverage of sensor points is visualized on a map in Fig. \ref{subfig:metrla_coverage}.

\subsubsection{Montreal Street-level Traffic Dataset} 
 We present the Montreal street-level traffic dataset (MSLTD), which provides traffic speed over 15-minute intervals \footnote{Publicly available at https://github.com/naghm3h/MSLTD}. The data was collected from street segments in Montreal, QC, Canada, during the six months from January 2019 to June 2019. The raw data is the historical travel time on road segments provided by the City of Montreal. The City of Montreal has deployed a network of sensors using Bluetooth technology on specific strategic road segments to present speed, travel time, and origin-destination information of journeys \cite{mtlOpendata} together with the information on the road segments where travel times are collected \cite{mtlOpendataSegment}. For the first six months of 2019, the data contains more than 14 million travel time records on 252 road segments. 

\begin{figure}[!t]
    \centering
    \begin{subfigure}{0.44\textwidth}
    \includegraphics[width=\textwidth]{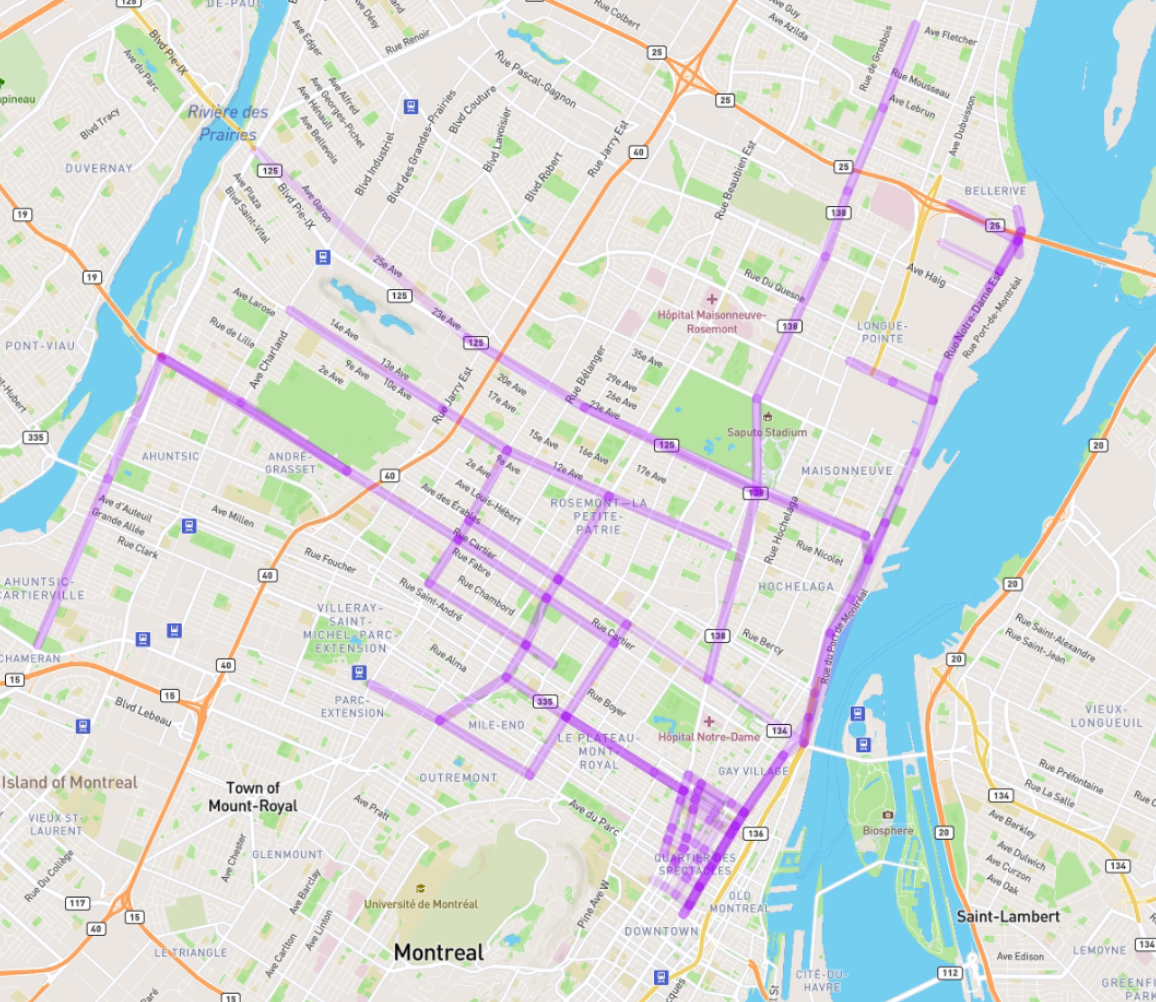}
            \caption{}
        \label{subfig:mtl_coverage}
    \end{subfigure}
    \begin{subfigure}{0.44\textwidth}    \includegraphics[width=\textwidth]{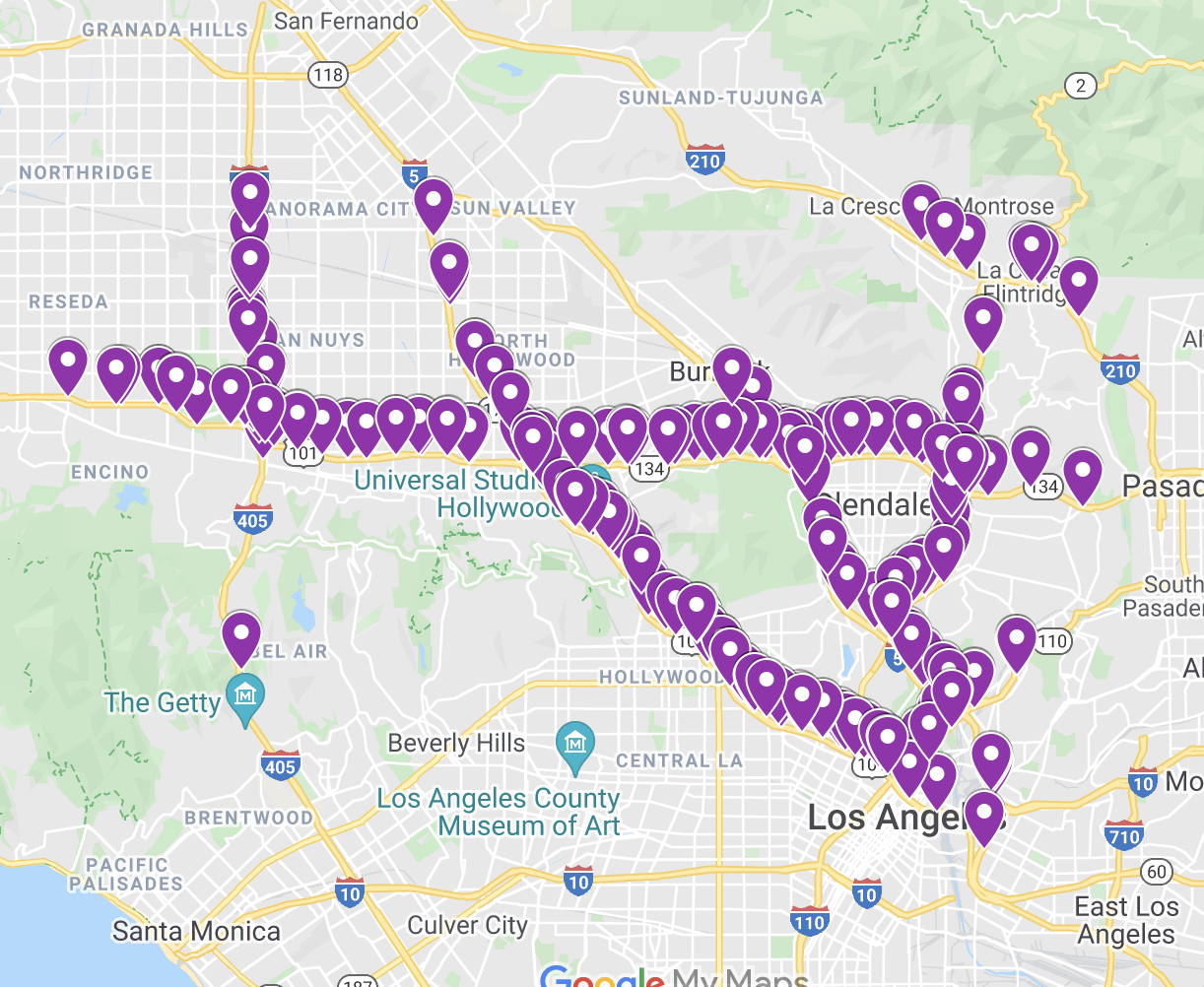}
            \caption{}
        \label{subfig:metrla_coverage}
    \end{subfigure}
    \caption{Datasets coverage (a) Montreal Street-level Traffic Dataset covering certain strategic road segments of Montreal. (b) METR-LA dataset sensor distribution.}
    \label{fig:maps2}
\end{figure}

The adjacency matrix corresponding to the pre-defined graph for GNN methods is calculated based on the road segment information. The variable features are mapped to the nodes, and the static states are considered as the edges to have a fixed graph structure. The goal is to capture network structural flow correlation since traffic speeds at one node should be correlated to traffic speeds at other nodes with which they share links. Each road segment is represented as a node and contains information including its origin and destination detector IDs. If two roads share the same start or end points, then a connection is formed in the graph.



\begin{center}
$\forall v_i, v_j | i\neq j, if OriginId_{i} == OriginId_{j} or $
$OriginId_{i} == DestinationId_{j} \Rightarrow e_{ij} \in \mathcal{E}$
\end{center}

To convert historical travel time data to traffic flow data, first, the travel time records whose corresponding segment information does not exist are removed. Since the dataset provides the average speed during a journey, for journeys taking more than 5 minutes, we split them into shorter time intervals.

\begin{figure}[!t]
    \centering
    \begin{subfigure}[t]{0.45\textwidth}
        \includegraphics[width=\textwidth]{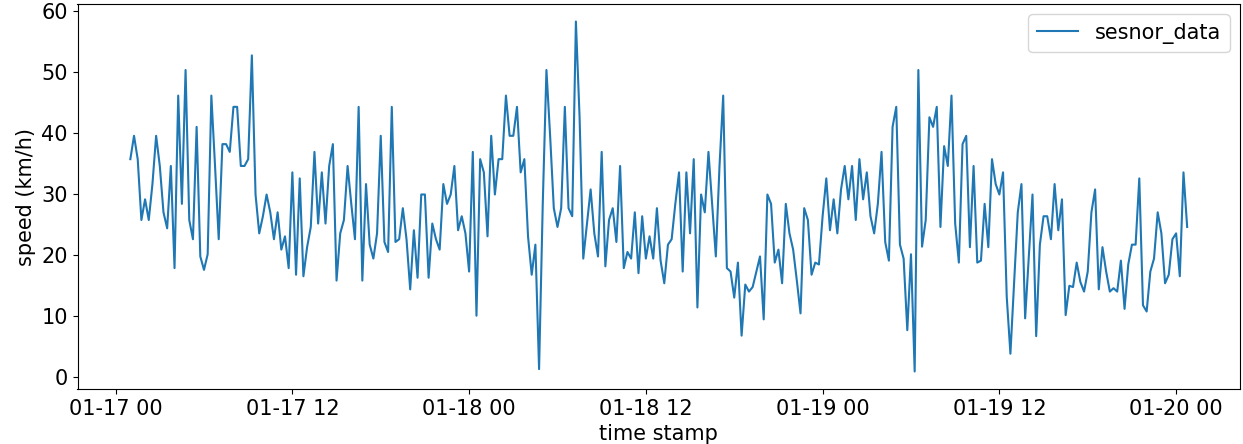}
        \caption{}
        \label{subfig:before_smth}
    \end{subfigure}
    
    \begin{subfigure}[t]{0.45\textwidth}
        \includegraphics[width=\textwidth]{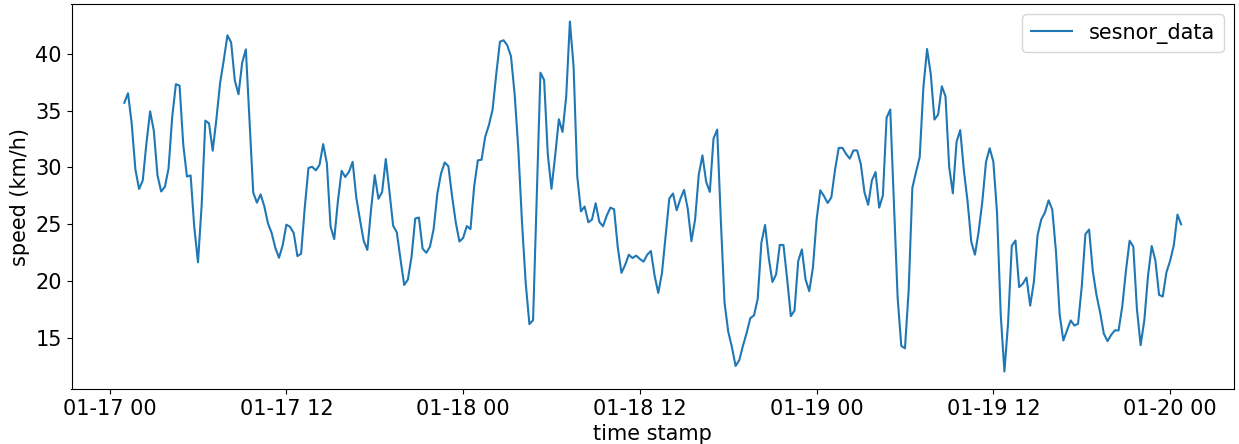}
        \caption{}
        \label{subfig:after_smth}
    \end{subfigure}
    \caption{Comparison of smoothing effect on data variation (a) before smoothing and (b) after smoothing operation using Gaussian filter.}
    \label{fig:smth}
\end{figure}

 \begin{algorithm}[H]
 \caption{Breaking up of long trips}
 \begin{algorithmic}[1]
 \renewcommand{\algorithmicrequire}{\textbf{Input:}}
 \REQUIRE $trips$ = $[\{link\_id, start\_time, end\_time\}]$
 \WHILE{$end\_time - start\_time > interval$}
 \STATE $trips.append(\{trip\_id,start\_time,start\_time+interval\})$
 \STATE $start\_time += interval$
 \ENDWHILE
 \RETURN $trips$
 \end{algorithmic}
 \end{algorithm}


Next, a temporal list is created for each link covering January 2019 to June 2019 at 5-minute intervals as shown in Fig. \ref{subfig:speed_matrix}. We fill in the list elements by the aggregated speed information from travel time records corresponding to each time interval as shown in Fig. \ref{subfig:links_table}, the filled elements of the table point to the multiple travel time record matching the time interval. The traffic state data (speed) is set to NAN for the elements whose time interval does not match any travel time record. Links containing more than $\tau$ consecutive NAN elements ($\tau = 1000$) are removed because of excessive missing data.

After removing rejected links, the adjacency matrix is revised to exclude those links. Ultimately, 105 major road segments with the highest number of trips are selected. Link coverage is shown in Fig. \ref{subfig:mtl_coverage}. At this step, the speed matrix is generated in 5-minute intervals for the selected segments and determines the speed through those specific segments. Since we are aggregating trips' data to fill in time intervals, for time intervals with multiple trips, we have multiple speed aggregation options as the representative speed of all trips in that particular timestamp, such as maximum, average, and minimum speeds of all trips that have occurred in that specific time range. In this work, for demonstrating MSLTD, we have selected the maximum speed since it better reflects the road network characteristics compared to other options that might be affected by drivers' behaviour, for example, the first element of the speed matrix in Fig. \ref{subfig:speed_matrix} is defined as:

\begin{center}
$1-1: max(22.15, 33.17, 2.61, 27.17, 34.01, 33.60, 55.97)$
\end{center}

for the element $1-2$ since there is no travel time record in its corresponding time interval, it is set to $NAN$. Element $3-3$ includes only one travel time record that would be the representative of this time interval itself.

Furthermore, for any remaining links with missing values, we aggregate the trip information in 15-minute time intervals to overcome the sparsity in the data. The rest of the missing temporal data for X-minute windows are replaced by the Historical Average method, using the weighted average of previous weeks.

The aggregation of the trip information may lead to noisy data for time intervals containing fewer trips, as shown in Fig. \ref{subfig:before_smth}. To address this, we apply a Gaussian filter to smooth the data, as shown in Fig. \ref{subfig:after_smth}.


Finally, the adjacency matrix represents road segment connectivity information, and the temporal matrix encodes the speed changes for all segments. 

\begin{table*}
\caption{Experimental results of various baseline methods conducted on METR-LA and Montreal datasets for different values of time steps ahead}
\label{performance_table}
\centering
\begin{tabular}{p{1.5cm}p{1.5cm}|p{1.5cm}|p{1.5cm}|p{1.5cm}|p{1.5cm}|p{1.5cm}|p{1.5cm}}
\hline
                       & Dataset & \multicolumn{3}{c|}{METR-LA}                                       & \multicolumn{3}{c}{MSLTD}                                                     \\ \hline
                       &         & \multicolumn{3}{c|}{Horizon}                                       & \multicolumn{3}{c}{Horizon}                                                      \\ \hline
\multicolumn{1}{l}{Method}                 & \multicolumn{1}{c}{Metric}  &  \multicolumn{1}{c}{30 min} &
\multicolumn{1}{c}{45 min} & \multicolumn{1}{c}{60 min} &  \multicolumn{1}{c}{30 min} &
\multicolumn{1}{c}{45 min} & \multicolumn{1}{c}{60 min}  \\ \hline

\multirow{3}{*}{HA} & \multicolumn{1}{c|}{RMSE}  & \multicolumn{1}{c}{7.77} & \multicolumn{1}{c}{7.77} & \multicolumn{1}{c|}{7.77} & \multicolumn{1}{c}{7.54} & \multicolumn{1}{c}{7.54} & \multicolumn{1}{c}{7.54} \\ & \multicolumn{1}{c|}{MAE}  & \multicolumn{1}{c}{4.15} & \multicolumn{1}{c}{4.15} & \multicolumn{1}{c|}{4.15} & \multicolumn{1}{c}{5.58} & \multicolumn{1}{c}{5.58} & \multicolumn{1}{c}{5.58} \\ & \multicolumn{1}{c|}{MAPE}  & \multicolumn{1}{c}{12.9\%} & \multicolumn{1}{c}{12.9\%} & \multicolumn{1}{c|}{12.9\%} & \multicolumn{1}{c}{28.20\%} & \multicolumn{1}{c}{28.20\%} & \multicolumn{1}{c}{28.20\%} \\ \hline

\multirow{3}{*}{VAR} & \multicolumn{1}{c|}{RMSE} & \multicolumn{1}{c}{9.37} & \multicolumn{1}{c}{10.01} & \multicolumn{1}{c|}{10.68} & \multicolumn{1}{c}{5.65} & \multicolumn{1}{c}{6.93} & \multicolumn{1}{c}{7.28} \\ & \multicolumn{1}{c|}{MAE} & \multicolumn{1}{c}{5.40} & \multicolumn{1}{c}{6.07} & \multicolumn{1}{c|}{6.50} & \multicolumn{1}{c}{4.27} & \multicolumn{1}{c}{5.25} & \multicolumn{1}{c}{5.51} \\ & \multicolumn{1}{c|}{MAPE} & \multicolumn{1}{c}{12.75\%} & \multicolumn{1}{c}{14.5\%} & \multicolumn{1}{c|}{15.84\%} & \multicolumn{1}{c}{19.85\%} & \multicolumn{1}{c}{25.46\%}   
& \multicolumn{1}{c}{27.09\%} \\ \hline

\multirow{3}{*}{GC-LSTM}  
& \multicolumn{1}{c|}{RMSE}  & \multicolumn{1}{c}{7.41}    
& \multicolumn{1}{c}{8.96}   & \multicolumn{1}{c|}{10.19}   
& \multicolumn{1}{c}{5.20}   & \multicolumn{1}{c}{6.39}   
& \multicolumn{1}{c}{7.13} \\ 
& \multicolumn{1}{c|}{MAE}  & \multicolumn{1}{c}{4.16}    
& \multicolumn{1}{c}{5.32}   & \multicolumn{1}{c|}{6.32}   
& \multicolumn{1}{c}{3.99}   & \multicolumn{1}{c}{4.77}   
& \multicolumn{1}{c}{5.22} \\ 
& \multicolumn{1}{c|}{MAPE}  & \multicolumn{1}{c}{11.38\%}    
& \multicolumn{1}{c}{13.75\%}   & \multicolumn{1}{c|}{15.68\%}   
& \multicolumn{1}{c}{19.62\%}   & \multicolumn{1}{c}{23.82\%}   
& \multicolumn{1}{c}{26.55\%} \\ \hline

\multirow{3}{*}{DCRNN}  
& \multicolumn{1}{c|}{RMSE}  & \multicolumn{1}{c}{6.45}    
& \multicolumn{1}{c}{7.21}   & \multicolumn{1}{c|}{7.59}   
& \multicolumn{1}{c}{5.13}   & \multicolumn{1}{c}{5.93}   
& \multicolumn{1}{c}{6.59} \\ 
& \multicolumn{1}{c|}{MAE}  & \multicolumn{1}{c}{3.15}    
& \multicolumn{1}{c}{3.42}   & \multicolumn{1}{c|}{3.60}   
& \multicolumn{1}{c}{3.63}   & \multicolumn{1}{c}{4.59}   
& \multicolumn{1}{c}{4.93} \\ 
& \multicolumn{1}{c|}{MAPE}  & \multicolumn{1}{c}{8.8\%}    
& \multicolumn{1}{c}{9.91\%}   & \multicolumn{1}{c|}{10.5\%}   
& \multicolumn{1}{c}{17.28\%}   & \multicolumn{1}{c}{21.42\%}   
& \multicolumn{1}{c}{23.05\%} \\ \hline

\multirow{3}{*}{MTGNN}  
& \multicolumn{1}{c|}{RMSE}  & \multicolumn{1}{c}{6.17}    
& \multicolumn{1}{c}{6.79}   & \multicolumn{1}{c|}{7.25}   
& \multicolumn{1}{c}{5.19}   & \multicolumn{1}{c}{6.53}   
& \multicolumn{1}{c}{6.89} \\ 
& \multicolumn{1}{c|}{MAE}  & \multicolumn{1}{c}{3.05}    
& \multicolumn{1}{c}{3.29}   & \multicolumn{1}{c|}{3.50}   
& \multicolumn{1}{c}{3.86}   & \multicolumn{1}{c}{4.84}   
& \multicolumn{1}{c}{5.10} \\ 
& \multicolumn{1}{c|}{MAPE}  & \multicolumn{1}{c}{8.21\%}    
& \multicolumn{1}{c}{8.94\%}   & \multicolumn{1}{c|}{9.88\%}   
& \multicolumn{1}{c}{17.46\%}   & \multicolumn{1}{c}{23.37\%}   
& \multicolumn{1}{c}{24.75\%} \\ \hline

\multirow{3}{*}{AGCRN}  
& \multicolumn{1}{c|}{RMSE}  & \multicolumn{1}{c}{9.96}    
& \multicolumn{1}{c}{11.52}   & \multicolumn{1}{c|}{12.72}   
& \multicolumn{1}{c}{5.27}   & \multicolumn{1}{c}{6.66}   
& \multicolumn{1}{c}{7.11} \\ 
& \multicolumn{1}{c|}{MAE}  & \multicolumn{1}{c}{4.31}    
& \multicolumn{1}{c}{5.06}   & \multicolumn{1}{c|}{5.74}   
& \multicolumn{1}{c}{3.93}   & \multicolumn{1}{c}{4.97}   
& \multicolumn{1}{c}{5.30} \\ 
& \multicolumn{1}{c|}{MAPE}  & \multicolumn{1}{c}{10.87\%}    
& \multicolumn{1}{c}{12.90\%}   & \multicolumn{1}{c|}{14.75\%}   
& \multicolumn{1}{c}{18.25\%}   & \multicolumn{1}{c}{24.18\%}   
& \multicolumn{1}{c}{26.18\%} \\ \hline

\multirow{3}{*}{MLS2S}  
& \multicolumn{1}{c|}{RMSE}  & \multicolumn{1}{c}{\bfseries 5.62}    
& \multicolumn{1}{c}{\bfseries 6.12}   & \multicolumn{1}{c|}{\bfseries 6.51}   
& \multicolumn{1}{c}{\bfseries 4.92}   & \multicolumn{1}{c}{\bfseries 5.86}   
& \multicolumn{1}{c}{\bfseries 6.28} \\ 
& \multicolumn{1}{c|}{MAE}  & \multicolumn{1}{c}{\bfseries 2.80}    
& \multicolumn{1}{c}{\bfseries 2.96}   & \multicolumn{1}{c|}{\bfseries 3.10}   
& \multicolumn{1}{c}{\bfseries 3.54}   & \multicolumn{1}{c}{\bfseries 4.27}   
& \multicolumn{1}{c}{\bfseries 4.57} \\ 
& \multicolumn{1}{c|}{MAPE}  & \multicolumn{1}{c}{\bfseries 7.33\%}    
& \multicolumn{1}{c}{\bfseries 7.96\%}   & \multicolumn{1}{c|}{\bfseries 8.48\%}   
& \multicolumn{1}{c}{\bfseries 16.31\%}   & \multicolumn{1}{c}{\bfseries 20.38\%}   
& \multicolumn{1}{c}{\bfseries 22.00\%} \\ \hline

\end{tabular}
\end{table*}



\subsection{Experimental Settings}
For both datasets, we use 70\% of the data for training, 10\% for
validation, and 20\% for testing. We use 12 historical time steps information to predict the next horizon steps. For 30-minutes to 1-hour prediction horizon varies between 6 and 12 on METR-LA data since it is provided in 5-minute time intervals. It varies between 2 to 4 for Montreal Street-level Traffic Dataset that is presented in 15 minutes intervals. We train the proposed model using the Adam optimizer with a base learning rate of 0.01 and decay ratio of 0.1. The network is implemented using PyTorch v1.7.1. All experiments are conducted on NVIDIA GeForce RTX 2080 Ti with 11GB memory. 

\subsection{Baseline Methods}
We compare MLS2S with the commonly used baseline methods and the state-of-the-art approaches: 1) Historical Average model (HA), which uses a weighted average of historical periods to predict future values; 2) Vector Auto-Regression (VAR), a statistical model of multivariate forecasting; 3) Diffusion Convolutional Recurrent Neural Network (DCRNN) \cite{li2017diffusion}, a graph-based method that  applies random walk on graph and uses common encoder-decoder structure, the spatial information in this method is provided to the network given a graph based on distances between sensors;  4) An improved version of Traffic Graph Convolutional Recurrent Neural Network (TGC-LSTM) \cite{cui2019traffic}, which defines the graph based on road network topology. In addition to the spatial information and temporal changes provided to the network, their model has an extra dependency on road network characteristics information, which is provided to the network as a free-flow reachability matrix. Following their network design, we achieved similar performance while removing this dependency, see experiment results in Appendix. Thus, in the experiments conducted, we use our modified version (GC-LSTM) of their approach by only providing spatial and temporal information to the network. 5) Adaptive Graph Convolutional Recurrent Network (AGCRN) \cite{zhou2017graph}, which considers temporal information as intra-dependencies and spatial information as inter-dependencies, then introduces two adaptive modules based on RNNs and GNNs, finally combines them to capture both dependencies; 6) Connecting the dots (MTGNN) \cite{wu2020connecting}, introduces a graph learning module to avoid using a predefined graph for multivariate time-series forecasting.

\subsection{Performance Comparison}
Table \ref{performance_table} shows a comparison of the performance for the above-mentioned methods. We employ three widely used measures in traffic forecasting literature to measure the performance, including (1) Mean Absolute Error (MAE), (2) Mean Absolute
Percentage Error (MAPE), and (3) Root Mean Squared
Error (RMSE).

\begin{equation}
	\label{eq:rmse_eval}
	RMSE = \frac{1}{n}\sum_{i=1}^{h}|\frac{Y^{t+i} - X^{t+i}}{Y^{t+i}}|
\end{equation}
\begin{equation}
	\label{eq:mape_eval}
	MAPE = \sqrt{\frac{1}{n} \sum_{i=1}^{h}(Y^{t+i} - X^{t+i})^2}
\end{equation}

The metrics are reported for three different prediction horizons, from 30 minutes to 1 hour. A comparison of the results performed on METR-LA dataset shows that the Historical Average method has a constant performance for different horizons since it relies on long-range information. Moreover, both baseline methods HA and VAR do not perform well compared to the deep neural network approaches. TGC-LSTM and AGCRN have a comparable performance which can be attributed to the fact that they both use GRU and GCN as their main network components. DCRNN achieves a better performance than the previously investigated methods, which shows the effectiveness of diffusion convolution operation in the encoder-decoder architecture. DCRNN and MTGNN performance results are almost in the same range for highway data. The proposed MLS2S approach outperforms all other methods on all reported metrics. In particular, the multilevel encoder design improves the MAE by more than 11\% for one-hour prediction.

DCRNN and MTGNN employ different mechanisms to learn spatial and temporal dependencies. While the MTGNN network avoids using predefined graphs, DCRNN initially needs spatial information to be provided to the network in a graph format. To learn temporal dependencies, DCRNN makes use of RNN family while MTGNN utilizes dilated convolution. Finally, comparing their performance on Montreal's urban road segments dataset clearly shows that DCRNN outperforms MTGNN specifically for greater horizon values. In MLS2S network design, we also use RNN for addressing temporal complexities and predefined graphs to provide spatial information to the network. Given the two top-performing methods on the urban road segments dataset (MLS2S and DCRNN), we can conclude that using a predefined graph is more effective for urban road connections with a higher density adjacency matrix. 

\begin{table}
\begin{center}
\caption{Efficiency comparison of two top performing methods}
\label{epoch_time}
\begin{tabular}{ c c | c | c }
\hline
 &  & \multicolumn{2}{c}{Training time per epoch(s)}\\ \hline
Method& & MSLTD & METR-LA\\
\hline
DCRNN& & 105 & 382\\ 
\hline
MLS2S& & 26.4 & 106\\ 
\hline
\end{tabular}
\end{center}
\end{table}

\subsection{Computation Time}
Table \ref{epoch_time} shows training time per epoch for DCRNN and MLS2S, which are the two top-performing methods based on forecasting performance that use a predefined graph in their structure. MSLTD generally runs faster than METR-LA since their dimensions differ. The number of nodes for the MSLTD dataset is smaller, and although it covers six months, it is provided in 15 minutes intervals. MLS2S method reduces time complexity by more than half on both MSLTD and METR-LA datasets. The DCRNN graph operation design adds to the number of graph operations by applying diffusion process and dual random walks, while in our sparse architecture design, we avoid these complexities, which significantly reduces the training time per epoch.

\section{Conclusion }
In this paper, we improve the efficiency and accuracy of graph-based deep-learning spatiotemporal forecasting with a novel multilevel sequence-to-sequence architecture for understanding correlations at different levels of abstraction. The proposed architecture is based on a sparse cell block that uses a traditional graph convolution operation combined with gated recurrent units to address the spatial and temporal dependencies. The cell block is designed in its simplest form to avoid complexities that result in additional computational time during the training of the network. To show the effectiveness of the proposed model on two different scales of traffic, we further introduced an urban street-level segment dataset collected from Montreal, QC, Canada over a 6-month time period. Experiments on two real-world datasets show the advantage of the proposed model in terms of computational time and performance improvement. 

Our work can inform decision-making for traffic management by providing improved predictions. On the other hand, the impact of computational time becomes more evident while running the model on a larger area with a considerable number of parameters in operational stages like predicting traffic covering all over the city.

As a spatiotemporal forecasting approach, this work has the potential to be applied to different research areas like meteorological forecasting, infectious disease surveillance, energy consumption, and economic analysis.

The road segment connections carry out more information than a pairwise relationship since all the links belong to the road network. For future work, we explore the higher-order relationship of data instead of the pairwise relationship represented by adjacent nodes information in graph-based neural network models.

{\appendix[GC-LSTM baseline method]

Cui et al. \cite{cui2019traffic} report the results of their experiments using the TGC-LSTM model on two traffic datasets. The first is proprietary and hence not available, and the second contains data gathered from four freeways of the Greater Seattle Area (LOOP Data) and is publicly available. We repeated the training under the same reported conditions and were able to replicate the reported results. When investigating the effect of the FFR module, we modified their original network by removing the FFR dependency. Table \ref{table:appx} shows the results of the comparison. 

\begin{table}[!htbp]
\begin{center}
\caption{Efficiency comparison of two top performing methods}
\label{table:appx}
\begin{tabular}{ c | c | c | c }
\hline
 & \multicolumn{3}{c}{Performance Metrics}\\ \hline
 Method & MAE & MAPE & RMSE\\
\hline
TGC-LSTM & 2.57 & 6.01 & 4.63\\ 
\hline
GC-LSTM & 2.56 & 6.12 & 3.65\\ 
\hline
\end{tabular}
\end{center}
\end{table}

The differences in the performance metrics between the TGC-LSTM \textit{with} and \textit{without} FFR is so negligible it does not justify the additional complexity of calculating and using the FFR. Based on this observation, and the fact that FFR is not readily available, we have used this modified version of TGC-LSTM as a baseline method, which is referred to as GC-LSTM. 

}

\bibliographystyle{IEEEtran}
\bibliography{main}

\end{document}